\documentclass[conference]{IEEEtran}
\IEEEoverridecommandlockouts
\usepackage{cite}
\usepackage{amsmath,amssymb,amsfonts}
\usepackage{algorithmic}
\usepackage{graphicx}
\usepackage{textcomp}
\usepackage{xcolor}
\usepackage{hyperref}
\def\BibTeX{{\rm B\kern-.05em{\sc i\kern-.025em b}\kern-.08em
    T\kern-.1667em\lower.7ex\hbox{E}\kern-.125emX}}
\begin{document}

\title{Transfer Learning and Organic Computing for Autonomous Vehicles}

\author{\IEEEauthorblockN{Christofer Fellicious}
\IEEEauthorblockA{\\
\textit{University of Passau}\\
Passau, Germany \\
fellic01@gw.uni-passau.de}
}
\maketitle

\begin{abstract}
Autonomous Vehicles(AV) are one of the brightest promises of the future which would help cut down fatalities and improve travel time while working in harmony. Autonomous vehicles will face with challenging situations and experiences not seen before. These experiences should be converted to knowledge and help the vehicle prepare better in the future. Online Transfer Learning will help transferring prior knowledge to a new task and also keep the knowledge updated as the task evolves.  This paper presents the different methods of transfer learning, online transfer learning and organic computing that could be adapted to the domain of autonomous vehicles.
\end{abstract}


\section{Introduction}
Autonomous Vehicles(AV) or Driver-less Cars are one of the most widely discussed emerging technology in the present day. An autonomous vehicle can be explained as a vehicle that adapts to its surroundings and can navigate itself by sensing its environment and minimal human input. The advantage of autonomous vehicles is that they will be able to reduce the number of injuries or even motor accidents in general, and also reduce the transit times. Most automobile companies like Volkswagen, Audi, BMW to tech companies like Google, nVidia, Tesla are engaging in research in autonomous vehicles. This shows that autonomous vehicles are the future of transportation.

The Society of Automotive Engineers(SAE) classifies autonomous vehicles into five different levels based of their capabilities with level 0 being simple warnings to the driver to level 5 where even the steering column is not required in the vehicle. While progressing through the stages, the cars accumulate years of shared experience of different unique types and scenarios that differ from region to region or even season to season. All this information from the sensors and the driver input should be processed into knowledge to be used to improve the autonomous functionalities of the vehicle. It is time consuming and inefficient to create models for specific tasks from scratch, whereas it is easier if the model adapts itself and helps apply the existing knowledge to a new task with minimal input. It is also impractical to rebuild a model every time an autonomous vehicle encounters a new scenario. Another issue for any machine learning task is the lack of availability of large annotated datasets of adequate quality to build highly precise models.

Transfer Learning is the method by which a preexisting predictor is adapted for newer tasks without completely retraining the predictor. There are many real world examples that can be seen among humans such as the ability to ice skate could be transferred to learn to in-line skate, or learning to ride a bicycle will help in learning to ride a motorcycle. "Organic computing aims at mastering complexity in technical systems by equipping technical systems with 'life-like' properties, i.e. by means of characteristics observed in natural systems"\cite{muller2017organic} such as self-learning and self-organization. Online Transfer Learning is a combination of transfer learning and the self-learning aspect of organic computing. The prior knowledge from an existing domain is transferred to a new domain to generate a new model, and then the model is continuously refined based on the self-learning ability of organic computing. Online Transfer Learning(OTL) helps build on the concept of transfer learning where the predictor observes only a few features at a time. The advantage of online transfer learning is that it enables the model to be updated continuously based on the arrival of new data.

Transfer Learning and Organic Computing are relevant in helping a vehicle learn and also assimilating the knowledge and extrapolating the existing knowledge to different tasks. The need for these methods is that it is virtually impossible to foresee all the possible permutations that autonomous vehicles will encounter during its use. The experiences could be from a tricky crossing to getting stuck in the sand. These methods will not only help solve new tasks but also keep fine tuning the model.

The contribution made by this paper is a survey of
\begin{itemize}
\item the different methods of transfer learning and online transfer learning that could be used for autonomous vehicles
\item the recent approaches in organic computing and how self-learning and self-organizing could be used in the domain of autonomous vehicles
\item how the different methods relate and complement each other to make autonomous vehicles more efficient
\end{itemize}

\section{Related Work}
A lot of research in recent years have been focused on the autonomous driving domain with nVidia releasing dedicated cards like the nVidia PX2 for the purpose. Zhou et al.\cite{rausch2017learning} has described an interesting method that shows how semantic features learned by Convolutional Neural Networks(CNN) can be transferred. They were motivated by the absence of perfectly annotated large datasets for the autonomous driving program. Pan et al.\cite{pan2017virtual} have shown that reinforcement learning and transfer learning can be used to train models for autonomous vehicles.

A comprehensive paper by janai et al.\cite{janai2017computer} reveals the problems faced by autonomous driving, the available datasets and the state-of-art methods. The paper provides an in-depth analysis of different methods used in autonomous vehicles such as pedestrian detection, optical flow, 3-D reconstruction, object recognition and segmentation etc. The paper highlights problems such as the limited availability of high quality optical flow datasets that directly affect the quality of the trained models.

Semantic Segmentation plays a very important role in understanding and interpreting a scene. Nigam et al.\cite{nigam2018ensemble} considers an ensemble model incorporating knowledge transfer based on drones for the semantic segmentation of aerial images.

The Passive Aggressive(PA) algorithm is popular for the online updation of models. Crammer et al.\cite{crammer2006online} discusses a family of algorithms for online learning predictions. The algorithm can be used from binary predictors to multi class predictors. The algorithm uses linear kernels but with the application of the kernel trick, it can be made to predict highly non-linear functions. 

\section{Methods of Learning}
This section explains the different methods in transfer learning and organic computing. 
\subsection{Transfer Learning}
Most machine learning algorithms assume that the training and testing data would come from the same feature space. In some situations, there might not be enough training data for a particular scenario. Consider that a vehicle needs to learn to drive in the mud. Instead of learning from scratch, it is possible that the vehicle can learn how to drive on the road using a large dataset and then fine tune that training by using a smaller dataset of driving in mud. The synopsis is that the car does not need to learn to drive from scratch, instead it could learn to drive and then adapt that knowledge to drive in mud. 
\begin{figure}
\includegraphics[width=\linewidth]{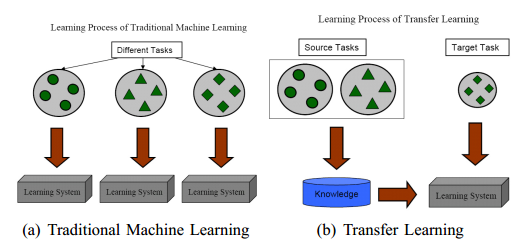}
\caption{Comparison between traditional and transfer learning}
\label{Fig:transfer_learning}
\end{figure}
Transfer learning relies on the fact that there might be basic similarities between the tasks, that can be carried over to the new one. This difference in traditional learning and transfer learning can be seen in the Fig \ref{Fig:transfer_learning}. Transfer Learning can be classified into two different sections based on the source domain and the target domain. They are
\begin{itemize}
\item Homogeneous Transfer Learning
\item Heterogeneous Transfer Learning
\end{itemize}

Homogeneous transfer learning takes place when the source and the target domains are present in the same feature space while in heterogeneous transfer learning they are in different feature spaces. Heterogeneous transfer learning is more difficult than the homogeneous mode because of the difference in feature spaces. "Most methods for heterogeneous transfer learning aim to learn a common feature
representation based on some correspondences between domains such that both source and target domain data can be
represented by homogeneous features"\cite{zhou2014hybrid}.

\subsubsection{Attention Mechanism based Transfer Learning}
Moon and Carbonell\cite{moon2017completely} proposes a novel approach where they assume that not all knowledge gained in the source domain can be transferred to the target domain. Their approach selectively transfers knowledge from the source domain to the target domain by using only a subset of the source knowledge and suppressing the rest which may have an adverse effect on the learning in the target domain. In this approach, a component known as the attention mechanism learns a set of parameters that contribute to a weight vector over a discrete subset of data. This weight vector describes the relevance of a particular feature vector for transferring knowledge. 

Their method first projects the source features and target features into a joint latent space via linear transformation. These features are then mapped into an embedded label space using a shared transformation.The authors use an autoencoder for generating the learned model. Since the source and target vectors share the same feature space, a parameter $\mu$ is used to penalize the attention mechanism and the joint space learning function if they learn only for the source domain. The advantage of this model is that by extracting only those parts of the source data that is beneficial for transfer, the model becomes better streamlined.

\subsubsection{Simultaneous multi task transfer learning}
Wong et al.\cite{wong2018transfer} proposes an automated transfer learning method where it is possible for the algorithm to learn several models suitable for different tasks. The authors assume that  most training tasks in neural networks have common design decisions such as network depth, training iterations, learning rate etc. The controller which is a Recurrent Neural Network(RNN), is capable of simultaneous multi-task training using two key features, (i) learned task representations and (ii) task-specific baseline and normalization. The initial feature consists of mapping each task to a unique embedding. The generation of network configurations is controlled by feeding the controller the embeddings at every time step, along with the action embeddings of the previous step. The controller then decides the appropriate action and passes it onwards to the Neural Network layers. The tasks are trained using policy gradient methods. Each task is also able to define a performance metric to be used as a reward. The reward affects the gradients applied to update the controller's policy for that specific task. The controller has to simultaneously keep track of multiple tasks and their rewards, and so it is necessary to ensure that the reward distributions have the same scale and variance. The authors claim that when a new task is given, the exploration can be significantly sped up by leveraging the learned biases about what combinations of parameter choices worked well together. The controller could learn an embedding for a new task and learn a representation that biases towards actions that performed well on other similar tasks. The advantage of this method is the ability to train multiple tasks simultaneously and the ability to establish a connection between a new task and a previously learned task. The downside to this algorithm is its memory intensive nature and that it will not perform well unless a suitably large number of tasks have been trained on it. 

\subsection{Organic Computing}
"Organic Computing is a research field emerging around the conviction that problems of organization in complex systems in computer science, telecommunications, neurobiology, molecular biology, ethology, and possibly even sociology can be tackled scientifically in a unified way. From the computer science point of view, the apparent ease in which living systems solve computationally difficult problems makes it inevitable to adopt strategies observed in nature for creating information processing machinery"\cite{OrganicComputing}. In other words, through organic computing researchers attempt to bring inherent strategies in nature such as self-learning and self-organizing  to machines. 

The organic computing machines form a sort of ensemble that could cooperate with each other and evolve over time. In such systems, each individual unit could be autonomous but when viewed as a whole, they could be seen as self organizing entity. An organic computing system will respond dynamically to changes in the environment and will also have the sufficient freedom to do so. When we consider connected autonomous vehicles as the basic entities, the application of organic computing methods and its benefits become clear. When considering autonomous vehicles of the future, it is expected that all vehicles will be interconnected and will communicate with each other to optimize travel times. The entire system can be seen as a sort of swarm with each vehicle having a distinct function for that particular time, i.e traveling from Point A to Point B. When considering such a swarm, it needs to self-organize itself dynamically, based on the inputs from individual components. This swarm can be seen as more than the sum of the individual elements. Suppose, there is heavy traffic on a particular path, the other vehicles should be rerouted. The whole system should self-organize in such a way that the adverse effects of any unintended scenario is minimized. Organic computing helps in this area of autonomous vehicles by making the whole network as a self-organizing mechanism.

Another application of organic computing is the self-learning paradigm. Once an entity, or in this case, an autonomous vehicle has learned a new skill or experience, the knowledge can be transfered to the other vehicles. When an autonomous vehicle is confronted with an entirely new task such as an unfamiliar terrain situation, the knowledge from a similar situation could be transferred for this particular task and then improved upon using organic computing. For example, trying to navigate a complex intersection where there are vehicles coming at varying speeds and densities.

\subsubsection{Self Learning with Dynamic Navigation Maps}
Lu et al.\cite{coombes7471503} demonstrates the self-learning capabilities of Unmanned Aerial Vehicle(UAV) for taxiing in the aerodromes. They consider the UAV to be present in a highly dynamic environment where there could be foreign objects, airport vehicles moving around or even maintenance work taking place. To handle such problems,they propose the concept of "dynamic navigation map", which is a collection of two maps, the aerodrome map and the obstacle map. The aerodrome map provides visual clues to what the camera expects to see, and the obstacle map is used to store the previously learned obstacle distribution. The obstacle map can be combined with the images from the sensor array to prepare a much more robust obstacle map. The self-learning part is handled by a Bayesian approach. The Bayesian inference is applied to the aerodrome map and the obstacle map separately. In this method, the probability of obstacles from the camera images are considered and their level of uncertainty is calculated. As the value of uncertainty increases, more relevance is placed on the aerodrome map which means that in such an uncertain situation more weight is given to prior knowledge. Using this method, if an object repeatedly appears in the images, it will be confirmed and learned with increasing confidence. The confidence measure is an advantage to the Bayesian method as the confidence measure cannot be obtained through normal Computer Vision algorithms.

\subsubsection{Optimizing vehicle behavior through reinforcement input}
M{\"u}ller-Schloer and Tomforde.\cite{muller2017organic} illustrates how order can be obtained as an effect of reinforcement. The author explains how the University of Michigan chose to put on concrete walkways in the most optimized paths between buildings. The University initially planted grass all over the campus and let the students choose their own path. After some time, the most frequently used trails would emerge where the grass will be almost non-existent. In this example, even with multiple options students would choose the most optimal path that would take them from building to building, and once the trails emerged it is reinforced as the students would continue to choose the most optimal trail. In the beginning, there maybe multiple trails emerging but only the most convenient are favored and reinforced. This could be used to solve issues with the autonomous vehicles as it progresses through the different levels of automation. Initially, the vehicle would be free to make any choice, but when an unfavorable choice is made the driver would intervene and rectify a mistake or the driver might take control in a previously unseen situation. In this context, we consider that the driver has prior knowledge and the choice made by the driver to be optimal. This intervention can be reinforced over time, changing the choice of the autonomous vehicle over time \cite{prakashincorporation}.

\subsection{Online Transfer Learning}
Online Transfer Learning is an extension of the Transfer Learning Framework and Organic Computing methods where the problem is addressed using an online learning framework. In an online learning framework, the algorithm observes instances in a sequential manner. After observing the instance, the algorithm makes a prediction based on its knowledge. The algorithm then might receive feedback indicating the correct output. Using the feedback it received, the algorithm may improve itself so that future classifications may have better accuracy.

\subsubsection{Ensemble based Online Transfer Learning}
Zhao and Hoi\cite{zhao2010otl} propose an ensemble based Online Transfer Learning(OTL) framework. The authors experiment with both homogeneous and heterogeneous data. In the homogeneous online transfer learning scenario, the authors first create a model based only on the target data. Later, an ensemble model is created which is a mixture of both the source and target data. A problem associated with  homogeneous transfer learning is that  the target variable to be predicted might change over time during training and this phenomenon is known as concept drift. In order to remove the problem of concept drift, the ensemble model features both predictors from source and target domains. The predictions of both the functions are combined using weights. There is a two step updating in the framework. Initially, the prediction function (f), updated by using online learning method which is the Passive Aggressive algorithm\cite{crammer2006online}. The second step is to update the prediction weights dynamically based on the current weights and a function of the square loss of the prediction.

In the heterogeneous online transfer learning experiment, Zhao and Hoi assume that the source data is a proper subset of the target data. The authors propose a multi-view approach for tackling the heterogeneous data problem as the source domain and target domain are very different. The authors also assume that the first m-dimensions of the target dataset features represents the source dataset features. Each data instance is split into two, where the first part represents the source domain and the second part represents the new target domain. This helps the ensemble classifiers to classify the new observed data sample correctly and forces the multi-view method not to deviate too much from the previous classifiers.

\subsubsection{Learning from multiple source domains}
Wu et al.\cite{wu7883886} describes an online transfer learning method where the knowledge from multiple source domains are considered and transferred to a single target domain. The authors assert that by building a model from multiple but related source domains for homogeneous transfer of knowledge the final model becomes more refined and learns to identify the core elements of knowledge that is to be transferred to the target domain. This approach is valid for the heterogeneous transfer learning too, where the authors assume that if the target domain feature space is split into two sets, the first set would contain homogeneous features which are shared by both the source and target domains while the second set would contain heterogeneous features. An image classification example stated by the authors is to obtain the subset of features of the target images from multiple source domains and then transfer the source knowledge to a new model. The authors combine multiple classifiers created from the source domains to form an ensemble classifier for the target domain. The source domain data is given in advance and so for each source domain a classifier is built in an offline learning paradigm. The target data is acquired in an online fashion and the Passive Aggressive algorithm is used for learning the representation of the target data. The loss for the decision is calculated by a hinge loss function and is added to the prior learned samples. 

In the scenario of heterogeneous transfer learning, the feature space is divided into two. The first part is assumed to be homogeneous with the source domain while the second part is heterogeneous with the source domain. A set of three base classifiers$(f^{s_{i}}, f^{T_{1}}_{i,t}, f^{T_{2}}_{i,t})$ are learned for each source domain", , where $f^{s_{i}}$ is the source domain classifier and $f^{T_{1}}_{i,t}, f^{T_{2}}_{i,t})$ are the target domain classifiers. $f^{T_{1}}_{i,t}$ and $f^{T_{2}}_{i,t}$ are learned by combining the first section and the second section in target domain with the source domain, respectively"\cite{wu7883886}. In the next step, the weights pertaining to each classifier is learned. The combination of the base classifiers and learned weights produces a robust classifier that can perform well in the target domain.

\subsubsection{Object tracking with Convolutional Neural Networks}
Wu et al.\cite{CHEN20161088} proposes a method "Online discriminative object tracking via deep convolutional neural network" where a neural network learns the discriminative features of an object and tracks the location and size of the object. The whole learning process consists of an initial transfer learning and the tracking is done as an online learning framework. The authors select a deep neural network because of the networks capability of learning high level feature representations of targets. The key idea of the authors for this experiment is to use the layers of the deep neural network as a generic and middle level image representation.
\begin{figure}
\includegraphics[width=\linewidth]{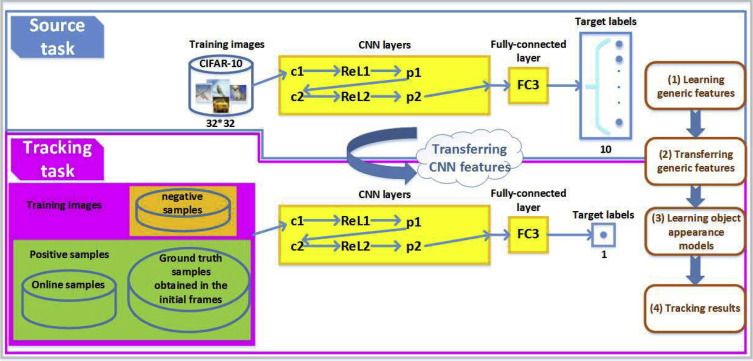}
\caption{Architecture of the Object Tracking Experiment\cite{CHEN20161088}}
\label{fig:neuralnetwork}
\end{figure}
The neural network is trained on the CIFAR-10\footnote{https://www.cs.toronto.edu/~kriz/cifar.html} dataset. Once the training is completed the parameters of all layers except the fully connected layer is transferred to the target task with only one output neuron. The network is then trained on the dataset of the object tracking task. The object tracking is done via a Particle Filtering Network that implements the Bayesian filter by Monte Carlo sampling. There are two main components to the particle filter, the dynamic model which generates candidate samples based on prior experience and the observation model that computes the similarity between the prediction and the actual value. For training, a manually annotated first image is given into the model, and then the coordinates of the object is obtained and a patch is extracted. The patch is rescaled to 32x32 pixel size(the size of the CIFAR-10 dataset images). Positive and negative samples are generated in by this method and the network is trained based on the generated data. 

A likelihood value is computed from the earlier trained neural network where the output neuron gives out a score. The likelihood is calculated by the equation
\begin{equation}
p(y_{t}|x_{t}) = exp(d_{t})
\label{Eq:1}
\end{equation}
While tracking an object, the appearance of the object might change due to its motion. So the likelihood function needs to adapt over time by fine tuning the neural network model. A main drawback of the appearance based model is its susceptance to drift, the model may slowly start to adapt to non targeted objects. To alleviate this problem, the authors accept the likelihood value only if it is above a certain threshold $T_{1}$ and the likelihood modifies the neural network only if the likelihood value is above a higher threshold $T_{2}$.  

\section{Conclusion and Discussion}
We outline methods in Transfer Learning, Organic Computing and online Transfer learning that can be used in the domain of autonomous vehicles. The paper shows how the different methods could work in tandem which would result in an output that would be better than what each individual component achieved. It is shown that online transfer learning is able to address the issues of limited availability of annotated data and the dynamically changing environments by self-learning. It is also possible that the models used in natural language processing can be extrapolated to the vision and audio domain easily. We believe online transfer learning would be one of the best options to look into for the development of autonomous vehicles. 

\bibliographystyle{IEEEtran}
\bibliography{IEEEexample}
\end{document}